\documentclass[11pt]{article}

\usepackage[final,nonatbib]{nips_2016}
\usepackage{comment}
\usepackage{color}
\usepackage{url}
\usepackage{listings}
\usepackage{pgfgantt}
\usepackage{xspace}
\usepackage{amsfonts, amssymb, amsmath, amsthm}
\usepackage{algorithm}
\usepackage{caption}
\usepackage{subcaption}
\usepackage{multirow}

\newcommand{\eat}[1]{}

\newif\ifsubmit
\submittrue

\ifsubmit
\newcommand{\chang}[1]{}
\newcommand{\dawn}[1]{}
\newcommand{\richard}[1]{}
\newcommand{\xinyun}[1]{}
\else
\newcommand{\chang}[1]{{\textcolor{blue}{[Chang: #1]}}}
\newcommand{\dawn}[1]{{\textcolor{blue}{[Dawn: #1]}}}
\newcommand{\richard}[1]{{\textcolor{blue}{[Richard: #1]}}}
\newcommand{\xinyun}[1]{{\textcolor{blue}{[Xinyun: #1]}}}
\fi

\newif\ifarxiv
\arxivtrue

\newcommand{\dam}{LA\xspace}
\newcommand{\NewAttention}{Latent Attention\xspace}

\begin{document}

\title{\NewAttention For If-Then Program Synthesis}
\author{Xinyun Chen\thanks{Part of the work was done while visiting UC Berkeley.}\\
  {\small Shanghai Jiao Tong University}
  \And
  Chang Liu\quad Richard Shin\quad Dawn Song\\
  UC Berkeley
  \And
  Mingcheng Chen\thanks{Work was done while visiting UC Berkeley. Mingcheng Chen is currently working at Google [X].}\\
  UIUC
}
\date{}
\maketitle

\begin{abstract}
  Automatic translation from natural language descriptions into programs is a long-standing
  challenging problem. In this work, we consider a simple yet important
  sub-problem: translation from textual descriptions to If-Then programs.
  We devise a novel neural network architecture for this task which we train end-to-end.
  Specifically, we introduce \NewAttention, which computes multiplicative weights for the words
  in the description in a two-stage process with the goal of better leveraging the natural language
  structures that indicate the relevant parts for predicting program elements.
  Our architecture reduces the error rate by $28.57\%$ compared to prior art~\cite{beltagy-quirk:2016:P16-1}.
  We also propose a one-shot learning scenario of
  If-Then program synthesis and simulate it with our existing dataset. We demonstrate a variation on the
  training procedure for this scenario that outperforms the original procedure, significantly
  closing the gap to the model trained with all data.
\end{abstract}

\section{Introduction}
\label{sec:intro}

A touchstone problem for computational linguistics is to translate natural
language descriptions into executable programs. Over the past decade, there
has been an increasing number of attempts to address this problem from both the natural
language processing community and the programming language community.
In this paper, we focus on a simple but important subset of programs containing
only one If-Then statement.

An If-Then program, which is also called a \emph{recipe}, specifies a \emph{trigger} and an \emph{action} function,
representing a program which will take the action when the trigger condition is met.
On websites, such as IFTTT.com, a user often
provides a natural language description of the recipe's functionality as well.
Recent work~\cite{quirk2015language,beltagy-quirk:2016:P16-1,dong2016language}
studied the problem of automatically synthesizing If-Then programs from their descriptions. In particular,
LSTM-based sequence-to-sequence approaches~\cite{dong2016language} and an approach
of ensembling a neural network and logistic regression~\cite{beltagy-quirk:2016:P16-1} were
proposed to deal with this problem.
In~\cite{beltagy-quirk:2016:P16-1}, however, the authors claim that the diversity of vocabulary
and sentence structures makes it difficult for an RNN to learn useful representations, and their ensemble
approach indeed shows better performance than the LSTM-based approach~\cite{dong2016language} on the
function prediction task (see Section~\ref{sec:background}).

In this paper, we introduce a new attention architecture, called \emph{\NewAttention},
to overcome this difficulty.
With \NewAttention, a weight is learned on each token to determine its importance for prediction of the trigger or the action.
Unlike standard attention methods, \NewAttention computes the token weights in a two-step process,
which aims to better capture the sentence structure.
We show that by employing \NewAttention over outputs of a bi-directional LSTM,
our new \NewAttention model can improve over the best prior
result~\cite{beltagy-quirk:2016:P16-1} by 5 percentage points from $82.5\%$ to
$87.5\%$ when predicting the trigger and action functions together, reducing
the error rate of~\cite{beltagy-quirk:2016:P16-1} by $28.57\%$.


Besides the If-Then program synthesis task proposed by~\cite{quirk2015language},
we are also interested in a new scenario. When a new trigger or action is released, the training data will contain few corresponding examples. We refer to this case as a \emph{one-shot learning} problem. We show that our \NewAttention model on top of dictionary embedding combining with a new training
algorithm can achieve a reasonably good performance for the one-shot learning task.

\section{If-Then Program Synthesis}
\label{sec:background}

\paragraph{If-Then Recipes.} In this work, we consider an important class of simple
programs called \textit{If-Then``recipes''} (or recipes for short), which are very
small programs for event-driven automation of tasks. Specifically, a recipe consists of a
trigger and an action, indicating that the action will be executed when the trigger is fulfilled.

The simplicity of If-Then recipes makes it a great tool for users who may not know how to code.
Even non-technical users can specify their goals using recipes,
instead of writing code in a more full-fledged programming language. A number of websites have embraced the If-Then programming paradigm and have
been hugely successful with tens of thousands of personal recipes created, including IFTTT.com and Zapier.com.
In this paper, we focus on data crawled from IFTTT.com.

IFTTT.com allows users to share their recipes publicly, along with short natural language
descriptions to explain the recipes' functionality. A recipe on IFTTT.com consists of a
\emph{trigger channel}, a \emph{trigger function}, an \emph{action channel}, an \emph{action function}, and arguments for the functions.
There are a wide range of channels, which can represent entities such as devices,
web applications, and IFTTT-provided services.
Each channel has a set of functions representing events (i.e., trigger functions)
or action executions (i.e., action functions).

For example, an IFTTT recipe with the following description
\[\text{Autosave your Instagram photos to Dropbox}\]
has the trigger channel {\tt Instagram}, trigger function {\tt Any\_new\_photo\_by\_you}, action channel {\tt Dropbox}, and action function
{\tt Add\_file\_from\_URL}. Some functions may take arguments. For example,
the {\tt Add\_file\_from\_URL} function takes three arguments:
the source URL, the name for the saved file, and the path to the destination folder.

\paragraph{Problem Setup.}
Our task is similar to that in~\cite{quirk2015language}. In particular,
for each description, we focus on predicting the channel and function for trigger and action respectively. Synthesizing a valid recipe also requires generating the arguments.
As argued by~\cite{beltagy-quirk:2016:P16-1}, however, the arguments are not crucial for
representing an If-Then program. Therefore, we defer our treatment for arguments
generation to
\ifarxiv
Appendix~\ref{app:arg},
\else
the supplementary material,
\fi
where we show that a simple frequency-based method can outperform all existing
approaches. In this way, our task turns into two classification problems for
predicting the trigger and action functions (or channels).

Besides the problem setup in~\cite{quirk2015language}, we also
introduce a new variation of the problem, a one-shot learning
scenario: when some new channels or functions are initially available,
there are very few recipes using these channels and functions in the
training set. We explore techniques to still achieve a reasonable
prediction accuracy on labels with very few training examples.

\section{Related Work}
\label{sec:work}

Recently there has been increasing interests in executable code generation. Existing works
have studied generating domain-specific code, such as regular expressions~\cite{kushman2013using},
code for parsing input documents~\cite{lei2013natural},
database queries~\cite{zelle1996learning, berant2013semantic}, commands to
robots~\cite{kate2005learning}, operating systems~\cite{branavan2009reinforcement},
smartphone automation~\cite{le2013smartsynth}, and spreadsheets~\cite{gulwani2014nlyze}.
A recent effort considers translating a mixed natural language and structured
specification into programming code~\cite{DBLP:journals/corr/LingGHKSWB16}.
Most of these approaches rely on semantic parsing~\cite{wong2006learning,
jones2012semantic,artzi2009broad,quirk2015language}. In particular,~\cite{quirk2015language}
introduces the problem of translating IFTTT descriptions into executable code,
and provides a semantic parsing-based approach. Two recent work studied approaches
using sequence-to-sequence model~\cite{dong2016language} and an ensemble of a neural
network and a logistic regression model~\cite{beltagy-quirk:2016:P16-1} to deal with
this problem, and showed better performance than~\cite{quirk2015language}. We show that
our \NewAttention method outperforms all prior approaches.
\xinyun{We cite the two ACL16 IFTTT papers without explanation here. Should we say something about them? (although they are briefly described in Section~\ref{sec:exp})}
Recurrent neural networks~\cite{zaremba2014recurrent, chung2014empirical}
along with attention~\cite{bahdanau2014neural} have demonstrated impressive results
on tasks such as machine translation~\cite{bahdanau2014neural},
generating image captions~\cite{xu2015show}, syntactic parsing~\cite{vinyals2015grammar}
and question answering~\cite{memn2n}.

\section{\NewAttention Model}
\label{sec:network}
\subsection{Motivation}

\newcommand{\trigger}[1]{{\color{blue}#1}}
\newcommand{\action}[1]{{\color{red}#1}}
\newcommand{\keyword}[1]{{\color{brown}#1}}

To translate a natural language description into a program, we would like to locate
the words in the description that are the most relevant for predicting desired labels
(trigger/action channels/functions).
For example, in the following description
\[\text{Autosave {\trigger{Instagram photos}} to your Dropbox folder}\]
the blue text ``Instagram photos'' is the most relevent for predicting the trigger.
To capture this information,
we can adapt the attention mechanism~\cite{bahdanau2014neural,memn2n}
---first compute a weight of the importance of each token in the sentence,
and then output a weighted sum of the embeddings of these tokens.

However, our intuition suggests that the weight for each token depends not only on the token itself, but also the overall sentence structure.
For example, in
\[\text{Post photos in your \trigger{Dropbox} folder \keyword{to} \action{Instagram}}\]
``Dropbox'' determines the trigger, even though in the previous example,
which contains almost the same set of tokens,
``Instagram'' should play this role. In this example, the prepositions such as ``to'' hint
that the trigger channel is specified in the middle of the description rather than at the end.
Taking this into account allows us to select ``Dropbox'' over ``Instagram''.

\NewAttention is designed to exploit such clues. We use the usual attention mechanism for computing a \emph{latent weight} for each token to determine which tokens in the sequence are more relevant to the trigger or the action.
These latent weights determine the final attention weights, which we call \emph{active weights}.
As an example, given the presence of the token ``to'', we might look at the tokens before ``to'' to determine the trigger.

\begin{figure}[t]
	\centering
	\includegraphics[scale=0.435]{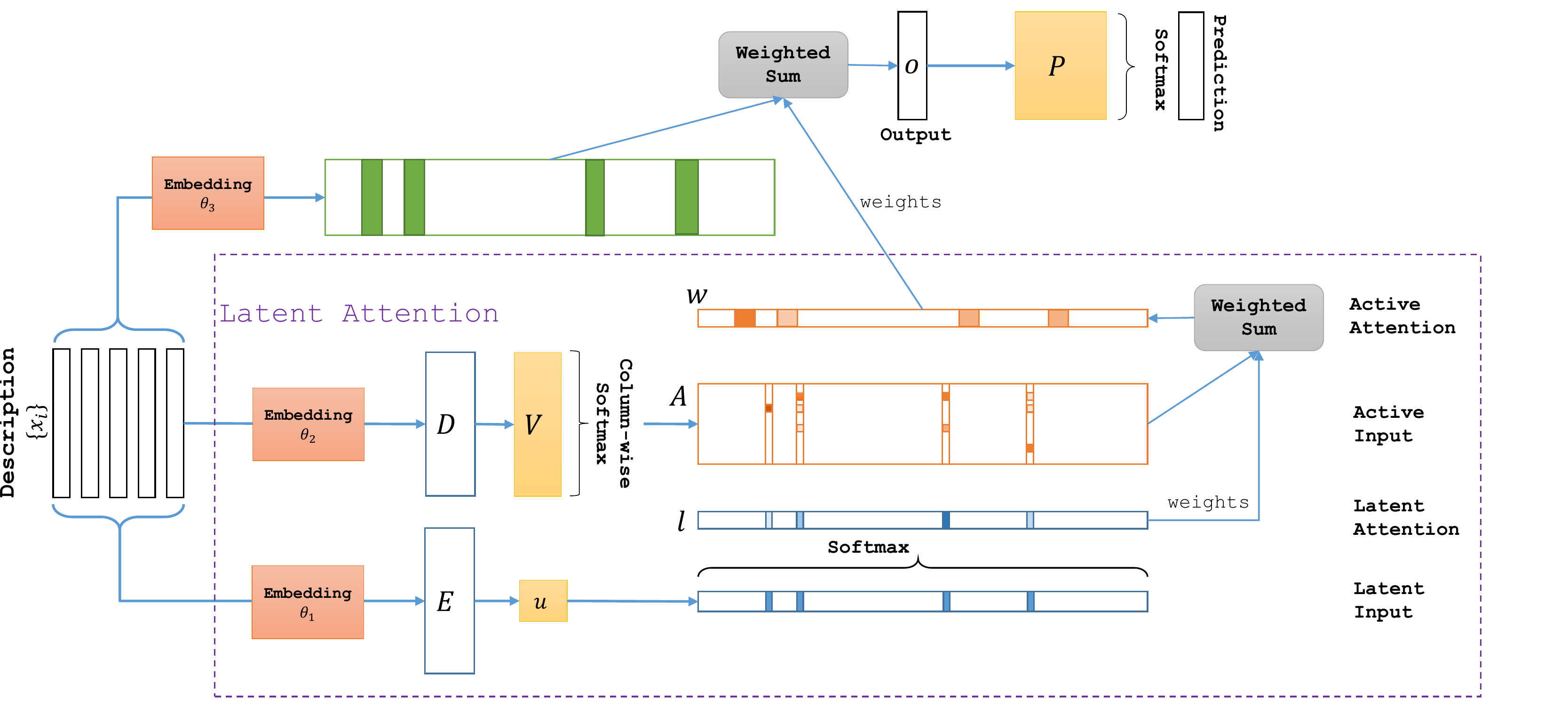}\\
	\caption{Network Architecture}\label{fig:arch}
\end{figure}

\subsection{The network}
The \NewAttention architecture is presented in Figure~\ref{fig:arch}. We follow
the convention of using lower-case letters to indicate column vectors, and capital letters for matrices.
Our model takes as input a sequence of symbols
$x_1,...,x_J$, with each coming from a dictionary of $N$ words.
We denote $X=[x_1,...,x_J]$. Here, $J$ is the maximal length of a description.
We illustrate each layer of the network below.

\newcommand{\emb}[2]{\ensuremath{\mathtt{Embed}_{#2}(#1)}}
\newcommand{\softmax}[1]{\ensuremath{\mathbf{softmax}(#1)}}

\paragraph{Latent attention layer.}  We assume each symbol $x_i$ is encoded
as a one-hot vector of $N$ dimensions. We can embed the input sequence $X$ into
a $d$-dimensional embedding sequence using $E=\emb{X}{\theta_1}$,
where $\theta_1$ is a set of parameters. We will discuss  different embedding methods in
Section~\ref{sec:details}. Here $E$ is of size $d\times J$.

The latent attention layer's output is computed as a standard softmax on top of
$E$. Specifically, assume that $l$ is the $J$-dimensional output vector,
$u$ is a $d$-dimensional trainable vector, we have
\[l = \mathbf{softmax}(u^T\emb{X}{\theta_1})\]

\paragraph{Active attention layer.} The active attention layer computes each token's weight
based on its importance for the final prediction. We call these weights \emph{active weights}.
We first embed $X$ into $D$ using another set of parameters $\theta_2$,
i.e., $D=\emb{X}{\theta_2}$ is of size $d\times J$. Next, for each token $D_i$, we compute its active
attention input $A_i$ through a softmax:
\[A_i = \softmax{VD_i}\]
Here, $A_i$ and $D_i$ denote the the $i$-th column vector of $A$ and $D$ respectively,
and $V$ is a trainable parameter matrix of size $J\times d$.
Notice that $VD_i=(VD)_i$, we can compute $A$ by performing
\emph{column-wise softmax} over $VD$. Here, $A$ is of size $J\times J$.

The active weights are computed as the sum of $A_i$, weighted by the output of latent attention
weight:~\xinyun{Here we call both $w$ and $A$ "active weights", which is a little confusing.
  Maybe it would be better to change the notation of current $A$ to some other letter,
  and not name it as "active weights", while change the notation of current $w$ into $A$, and call it "active weights".}
\[w = \sum_{i=1}^{J}l_iA_i = Al\]

\paragraph{Output representation.} We use a third set of parameters $\theta_3$ to embed $X$ into a
$d\times J$ embedding matrix, and the final output $o$, a $d$-dimensional vector, is the sum of the embedding weighted by the active weights:
\[o=\emb{X}{\theta_3}w\]

\paragraph{Prediction.} We use a softmax to make the final prediction: $\hat{f}=\mathbf{softmax}(Po)$, where $P$ is a $d\times M$ parameter matrix, and $M$ is
the number of classes.

\subsection{Details}
\label{sec:details}


\paragraph{Embeddings.}
We consider two embedding methods for representing words in the vector space.
The first is a straightforward word embedding, i.e., $\emb{X}{\theta}=\theta X$, where $\theta$
is a $d\times N$ matrix and the rows of $X$ are one-hot vectors
over the vocabulary of size $N$.
We refer to this as ``dictionary embedding" later in the paper.
$\theta$ is not pretrained with a different dataset or objective, but initialized randomly and learned
at the same time as all other parameters.
We observe that when using \NewAttention, this simple method is effective enough to outperform some recent results
~\cite{quirk2015language, dong2016language}.

The other approach is to take the word embeddings, run them through a bi-directional LSTM (BDLSTM)~\cite{zaremba2014recurrent}, and
then use the concatenation of two LSTMs' outputs at each time step as the embedding.
This can take into account the context around a token, and thus the embeddings should contain
more information from the sequence than from a single token. We refer to such
an approach as ``BDLSTM embedding". The details are deferred to
\ifarxiv
Appendix~\ref{app:arch}.
\else
the supplementary material.
\fi
In our experiments, we observe that with the help of this embedding method, \NewAttention can outperform the prior state-of-the-art.

In \NewAttention, we have three sets of embedding parameters, i.e., $\theta_1, \theta_2,\theta_3$. In practice, we find that we
can equalize the three without loss of performance. Later, we will show that keeping them separate is helpful for our one-shot learning setting.

\paragraph{Normalizing active weights.} We find that normalizing the active weights $a$
before computing the output is helpful to improve the performance. Specifically, we compute
the output as
\[o=\emb{X}{\theta}\mathbf{normalized}(w)=\emb{X}{\theta}\frac{w}{||w||}\]
where $||w||$ is the $L_2$-norm of $w$. In our experiments, we observe that this normalization can improve the
performance by 1 to 2 points.

\paragraph{Padding and clipping.} \NewAttention requires a fixed-length input sequence.
To handle inputs of variable lengths, we perform padding and clipping. If an input's length
is smaller than $J$, then we pad it with null tokens at the end of the sequence.
If an input's length
is greater than $J$ (which is 25 in our experiements), we keep the first 12 and the last 13 tokens, and get rid of all the rest.

\newcommand{\unktok}{$\langle$UNK$\rangle$}

\paragraph{Vocabulary.} We tokenize each sentence by splitting on whitespace and punctuation
(e.g., \texttt{$.,!?"':;)($} ), and convert all characters into lowercase. We keep all punctuation
symbols as tokens too. We map each of
the top 4,000 most frequent tokens into themselves, and all the rest into a special
token \unktok. Therefore our vocabulary size is 4,001.
Our implementation has no special handling for typos.

\section{If-Then Program Synthesis Task Evaluation}
\label{sec:exp}

In this section, we evaluate our approaches with
several baselines and previous work~\cite{quirk2015language,
beltagy-quirk:2016:P16-1,dong2016language}.
We use the same crawler from Quirk et al.~\cite{quirk2015language} to crawl
recipes from IFTTT.com. Unfortunately, many recipes are no longer available.
We crawled all remaining recipes, ultimately obtaining 68,083 recipes for the training set.
\cite{quirk2015language} also provides a list of 5,171 recipes for validation,
and 4,294 recipes for test. All test recipes come with labels from Amazon
Mechanical Turk workers. We found that only 4,220 validation recipes and
3,868 test recipes remain available.
\cite{quirk2015language} defines a subset of test recipes, where each recipe has at least 3
workers agreeing on its labels from IFTTT.com, as the gold testset. We find that
584 out of the 758 gold test recipes used in~\cite{quirk2015language} remain available.
We refer to these recipes as the \emph{gold test set}.
We present the data statistics
\ifarxiv
in Appendix~\ref{app:data:stat}.
\else
in the supplementary material.
\fi

\paragraph{Evaluated methods.} We evaluate two embedding methods as well as the
effectiveness of different attention mechanisms. In particular, we compare no attention,
standard attention, and \NewAttention. Therefore, we evaluate six architectures in total.
When using dictionary embedding with no attention, for each sentence, we sum the embedding of each word, then pass it through a softmax
layer for prediction. For convenience, we refer to such a process as \emph{standard softmax}. For BDLSTM with no attention, we concatenate final states of forward and backward LSTMs, then pass the concatenation through a softmax layer for prediction.
The two embedding methods with standard
attention mechanism~\cite{memn2n} are described in
\ifarxiv
Appendix~\ref{app:arch}.
\else
the supplementary material.
\fi
The \NewAttention models have been presented in Section~\ref{sec:network}.

\paragraph{Training details.}
For architectures with no attention, they were trained using a learning rate of
0.01 initially, which is multiplied by 0.9 every 1,000 time steps. Gradients with
$L_2$ norm greater than 5 were scaled down to have norm 5. For architectures
with either standard attention mechanism or \NewAttention, they were trained using a
learning rate of 0.001 without decay, and gradients with $L_2$ norm greater than 40 were scaled down to have norm 40. All models were trained using Adam~\cite{kingma2014adam}. All weights were initialized uniformly randomly in $[-0.1, 0.1]$.
Mini-batches were randomly shuffled during training. The mini-batch size is 32 and the embedding vector size $d$ is 50.

\begin{figure}[t]
	\begin{minipage}{0.5\linewidth}
		\includegraphics[scale=0.17]{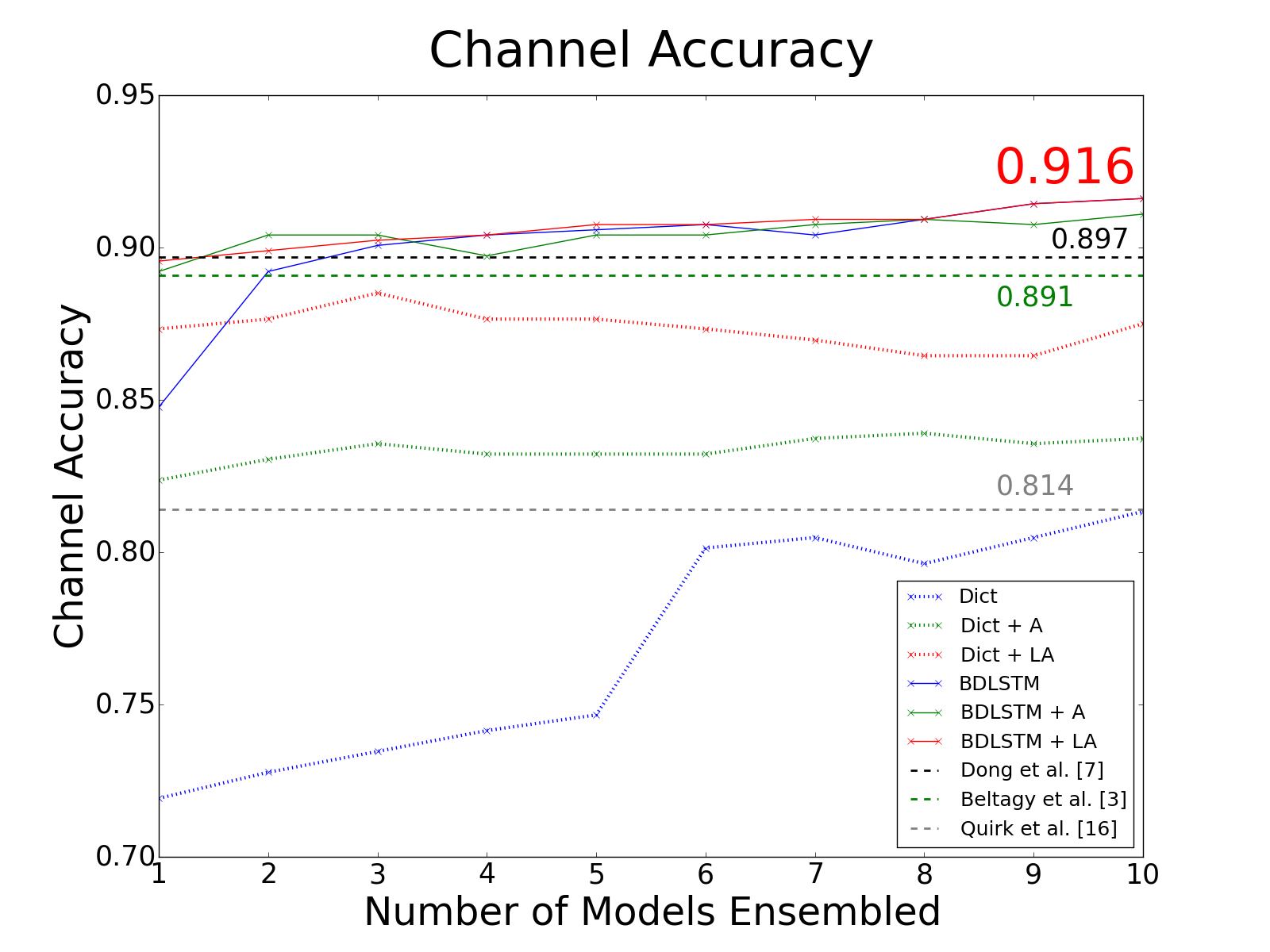}
		\caption{Accuracy for Channel}
		\label{fig:acc-channel}
	\end{minipage}
	\begin{minipage}{0.45\linewidth}
		\includegraphics[scale=0.17]{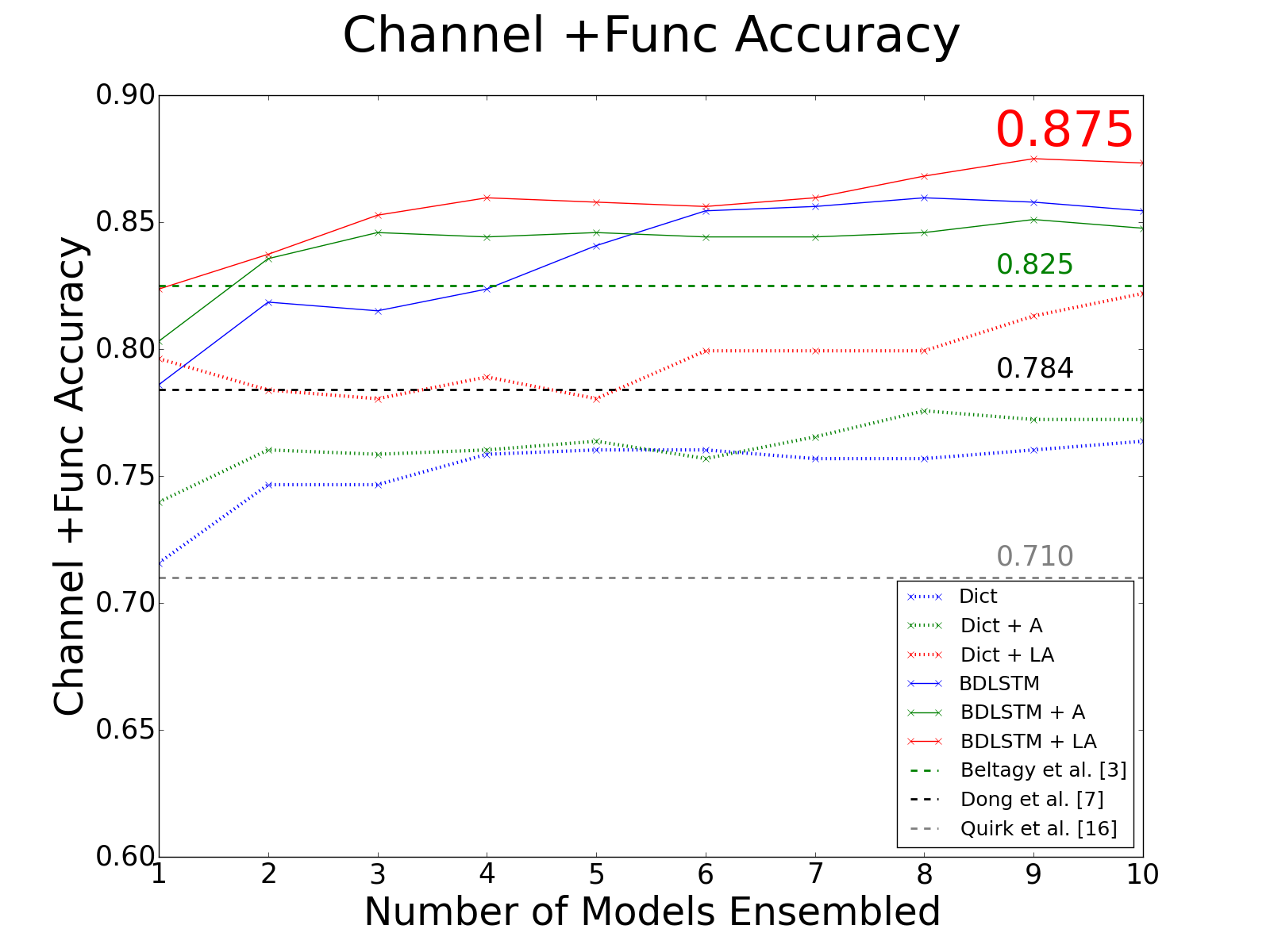}
		\caption{Accuracy for Channel+Function}
		\label{fig:acc-func}
	\end{minipage}
\end{figure}

\paragraph{Results.}
Figure~\ref{fig:acc-channel} and Figure~\ref{fig:acc-func} present the
results of prediction accuracy on channel and function respectively.
Three previous works' results are presented as well. In particular,
\cite{quirk2015language} is the first work introducing the If-Then program synthesis
task. \cite{dong2016language} investigates the approaches using
sequence-to-sequence models, while \cite{beltagy-quirk:2016:P16-1} proposes an approach
to ensemble a feed-forward neural network and a logistic regression model.
The numerical values for all data points can be found in
\ifarxiv
Appendix~\ref{app:data:stat}.
\else
the supplementary material.
\fi

For our six architectures, we use 10 different random initializations to train 10
different models. To ensemble $k$ models, we choose the best $k$ models on the validation set
among the 10 models, and average their softmax outputs as the ensembled output.
For the three existing approaches~\cite{quirk2015language, dong2016language,beltagy-quirk:2016:P16-1}, we choose the best results
from these papers.

We train the model to optimize for function prediction accuracy. The channel accuracy in Figure~\ref{fig:acc-channel} is
computed in the following way: to predict the channel, we first predict the function (from a list of all functions in all channels),
and the channel that the function belongs to is returned as the predicted channel.
We observe that
\begin{itemize}
	\item \NewAttention steadily improves over standard attention architectures and no attention ones using either embedding method.
	\item In our six evaluated architectures, ensembling improves upon using only one model significantly.
    \item When ensembling more than one model, BDLSTM embeddings perform better than dictionary embeddings.
        We attribute this to that for each token, BDLSTM can encode the information of its surrounding tokens, e.g., phrases, into its embedding, which
        is thus more effective. 
      \item For the channel prediction task in Figure~\ref{fig:acc-channel}, all architectures except dictionary
        embedding with no attention (i.e., Dict) can
        outperform~\cite{quirk2015language}. Ensembling only 2 BDLSTM models with either standard attention
        or \NewAttention
        is enough to achieve better performance than prior art~\cite{dong2016language}.
        By ensembling 10 BDLSTM+\dam models, we can improve
        the latest results~\cite{dong2016language} and~\cite{beltagy-quirk:2016:P16-1}
        by 1.9 points and 2.5 point respectively.
    \item For the function prediction task in Figure~\ref{fig:acc-func}, all our six models (including Dict) outperform~\cite{quirk2015language}. Further, ensembling 9 BDLSTM+\dam can improve the previous best results~\cite{beltagy-quirk:2016:P16-1}
        by 5 points. In other words, our approach reduces the error rate of~\cite{beltagy-quirk:2016:P16-1} by 28.57$\%$.
\end{itemize}

\section{One-Shot Learning}
\label{sec:osl}

We consider the scenario when websites such as IFTTT.com
release new channels and functions. In such a scenario, for a period of time, there
will be very few recipes using the newly available channels and fucntions; however,
we would still like to enable synthesizing If-Then programs using these new functions.
The rarity of such recipes in the training set creates a challenge
similar to the \textit{one-shot learning} setting.
In this scenario, we want to leverage the large amount of recipes for existing functions, and the goal is to achieve a good prediction accuracy for the new functions without significantly compromising the overall accuracy.

\subsection{Datasets to simulate one-shot learning}
\label{subsec:osl-dataset}
To simulate this scenario with our existing dataset, we build two one-shot variants of it as follows.
We first split the set of trigger functions into
two sets, based on their frequency. The top100 set contains
the top 100 most frequently used trigger functions, while the non-top100 set contains the rest.

Given a set of trigger functions $S$, we can build a skewed training set to include
all recipes using functions in $S$, and 10 randomly chosen recipes for each function not in $S$.
We denote this skewed training set created based on $S$ as $(S, \overline{S})$, and refer to
functions in $S$ as \emph{majority functions} and functions in $\overline{S}$ as \emph{minority functions}.
In our experiments, we construct two new training sets by choosing $S$ to be the top100 set
and non-top100 set respectively. We refer to these two training sets as SkewTop100 and SkewNonTop100.

The motivation for creating these datasets is to mimic two different scenarios. On one hand, SkewTop100 simulates
the case that at the startup phase of a service, popular recipes are first published, while less frequently used
recipes are introduced later. On the other hand, SkewNonTop100 captures the
opposite situation.
The statistics for these two training sets are presented in
\ifarxiv
Appendix~\ref{app:data:stat}.
\else
the supplementary material.
\fi
While SkewTop100 is more common in real life, the SkewNonTop100 training set is only $15.73\%$ of
the entire training set, and thus is more challenging.

\subsection{Training}

We evaluate three training methods as follows, where the last one is specifically designed for attention mechanisms.
In all methods, the training data is either SkewTop100 or SkewNonTop100.
	\vspace{-2em}
	\paragraph{Standard training.} We do not modify the training process.
	\vspace{-0.7em}
	\paragraph{Na\"{i}ve two-step training.} We do standard training first.
	Since the data is heavily skewed, the model may behave poorly on the minority functions.
	From a training set $(S, \overline{S})$, we create a rebalanced dataset,
	by randomly choosing 10 recipes for each function in $S$ and all recipes using functions in $\overline{S}$.
    Therefore, the numbers of recipes using each function are similar in this rebalanced dataset. We recommence the training using this rebalanced training dataset in the second step.
	\vspace{-0.7em}
	\paragraph{Two-step training.} We still do standard training first, and
	then create the rebalanced dataset in the similar way as that in na\"{i}ve two-step training. However, in the second step, instead of training the entire network,
	we keep the attention parameters fixed, and train only the parameters in the
    remaining part of the model. Take the \NewAttention model depicted in Figure~\ref{fig:arch} as an example. In the second step, we keep parameters $\theta_1$, $\theta_2$, $u$, and $V$ fixed, and only
  update $\theta_3$ and $P$ while training on the rebalanced dataset.
	We based this procedure on the intuition that since the rebalanced dataset is very small,
  fewer trainable parameters enable easier training.
	

\begin{figure}[t]
\centering
		\begin{subfigure}{0.48\linewidth}
			\includegraphics[scale=0.5]{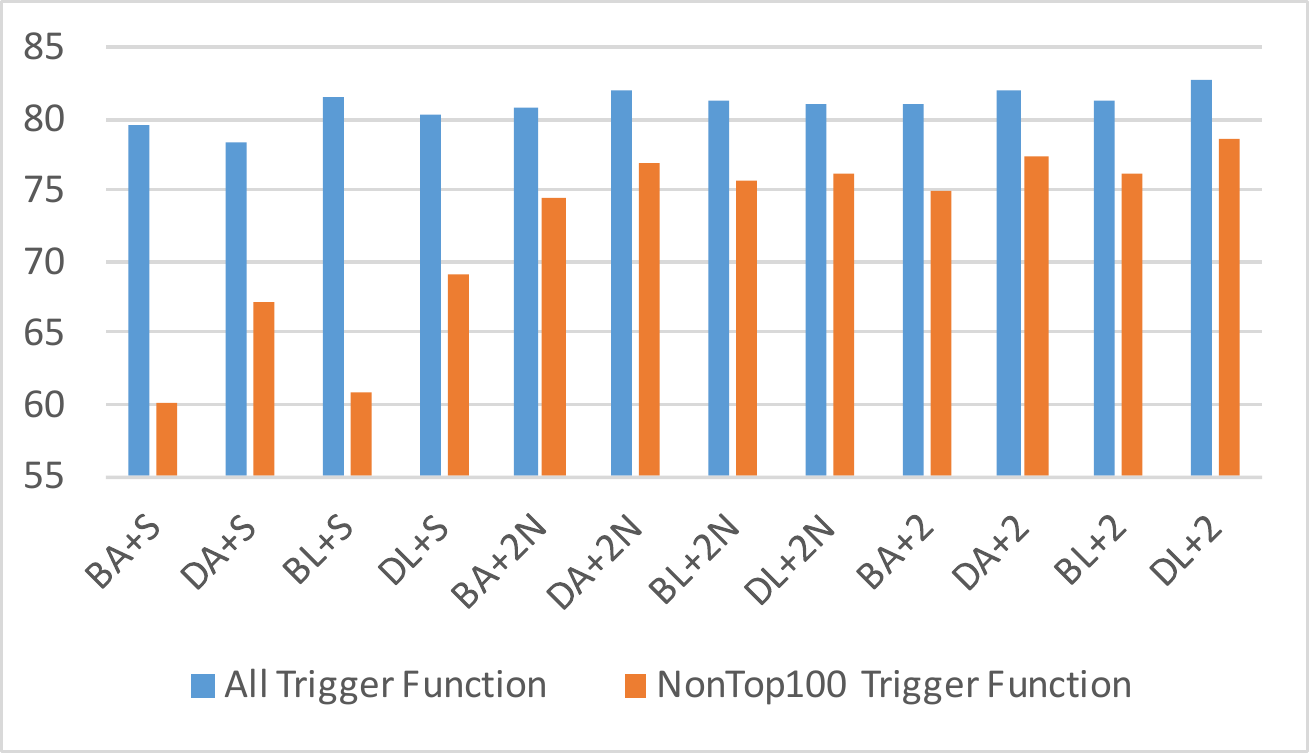}
			\caption{Trigger Function Accuracy (SkewTop100)}
			\label{fig:top100}
		\end{subfigure}
		\begin{subfigure}{0.45\linewidth}
			\includegraphics[scale=0.5]{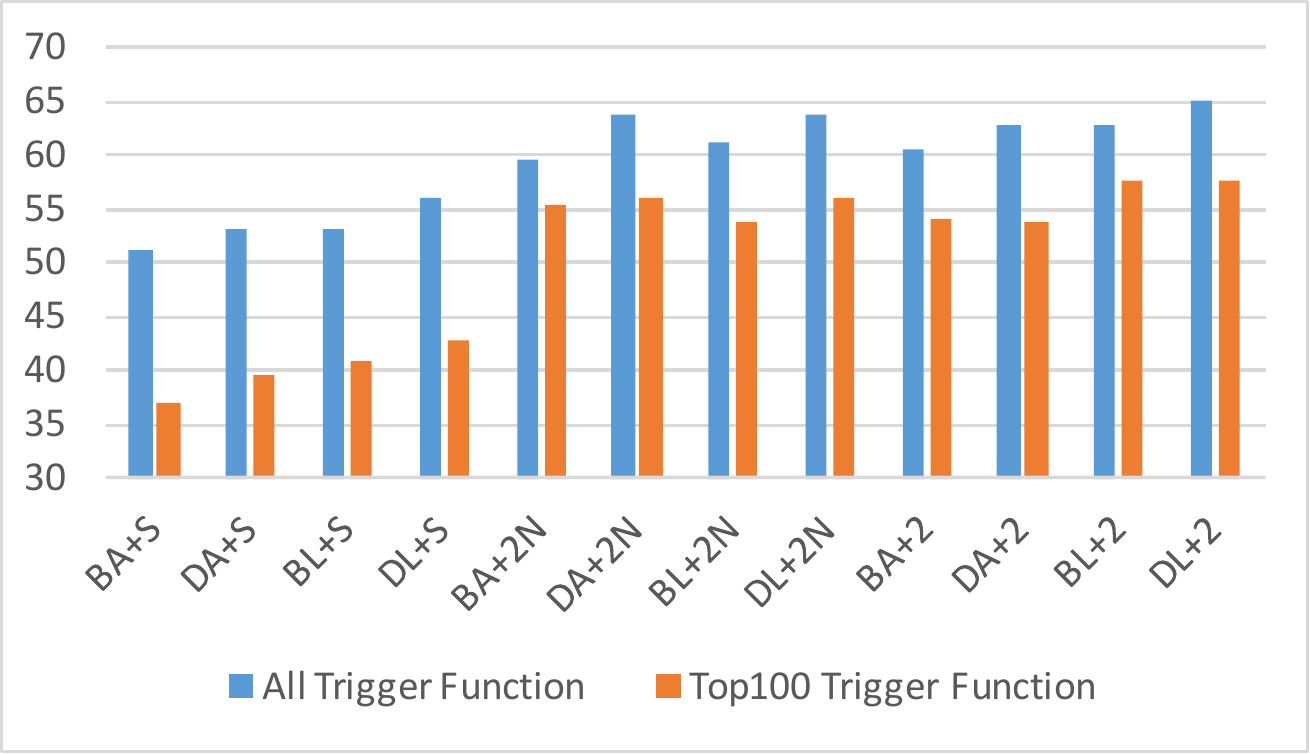}
			\caption{Trigger Function Accuracy (SkewNonTop100)}
			\label{fig:nontop100}
		\end{subfigure}
	\caption{One-shot learning experiments. For each column XY-Z, X from \{B, D\} represents whether
            the embedding is BDLSTM or Dictionary; Y is either empty, or is from \{A, L\}, meaning that either no
            attention is used, or standard attention or \NewAttention is used; and Z is from \{S, 2N, 2\},
            denoting standard training, na\"{i}ve two-step training or two-step
            training.
            }
	\label{fig:one-shot}
\end{figure}

\subsection{Results}

We compare the three training strategies using our proposed models.
We omit the no attention models, which do not perform better than attention models
and cannot be trained using two-step training.
We only train one model per strategy, so the results are without ensembling.
The results are presented in Figure~\ref{fig:one-shot}. The concrete values can be found in
\ifarxiv
Appendix~\ref{app:data:stat}.
\else
the supplementary material.
\fi
For reference, the best single BDLSTM+\dam model can achieve
$89.38\%$ trigger function accuracy: $91.11\%$ on top100 functions, and $85.12\%$ on non-top100 functions.
We observe that
\begin{itemize}
	\item Using two-step training, both the overall accuracy and
		the accuracy on the minority functions are generally
        better than using standard training and na\"{i}ve two-step training.
	
	\item \NewAttention outperforms standard attention when using the same training method.
	
	\item The best \NewAttention model (Dict+\dam) with two-step training can achieve
	$82.71\%$ and $64.84\%$ accuracy for trigger function on the gold test set, when trained on the SkewTop100 and SkewNonTop100 datasets respectively.
  For comparison, when using the entire training dataset, trigger function accuracy of Dict+\dam is $89.38\%$.
  Note that the SkewNonTop100 dataset accounts for only $15.73\%$ of the entire training dataset.
	
	\item For SkewTop100 training set, Dict+\dam model can achieve $78.57\%$ accuracy on
	minority functions in gold test set. This number for using the full training dataset is $85.12\%$,
    although the non-top100 recipes in SkewTop100 make up only $30.54\%$ of those in the full training set.
\end{itemize}

\section{Empirical Analysis of \NewAttention}

\begin{figure}[t]
  \centering
  \includegraphics[width=\linewidth]{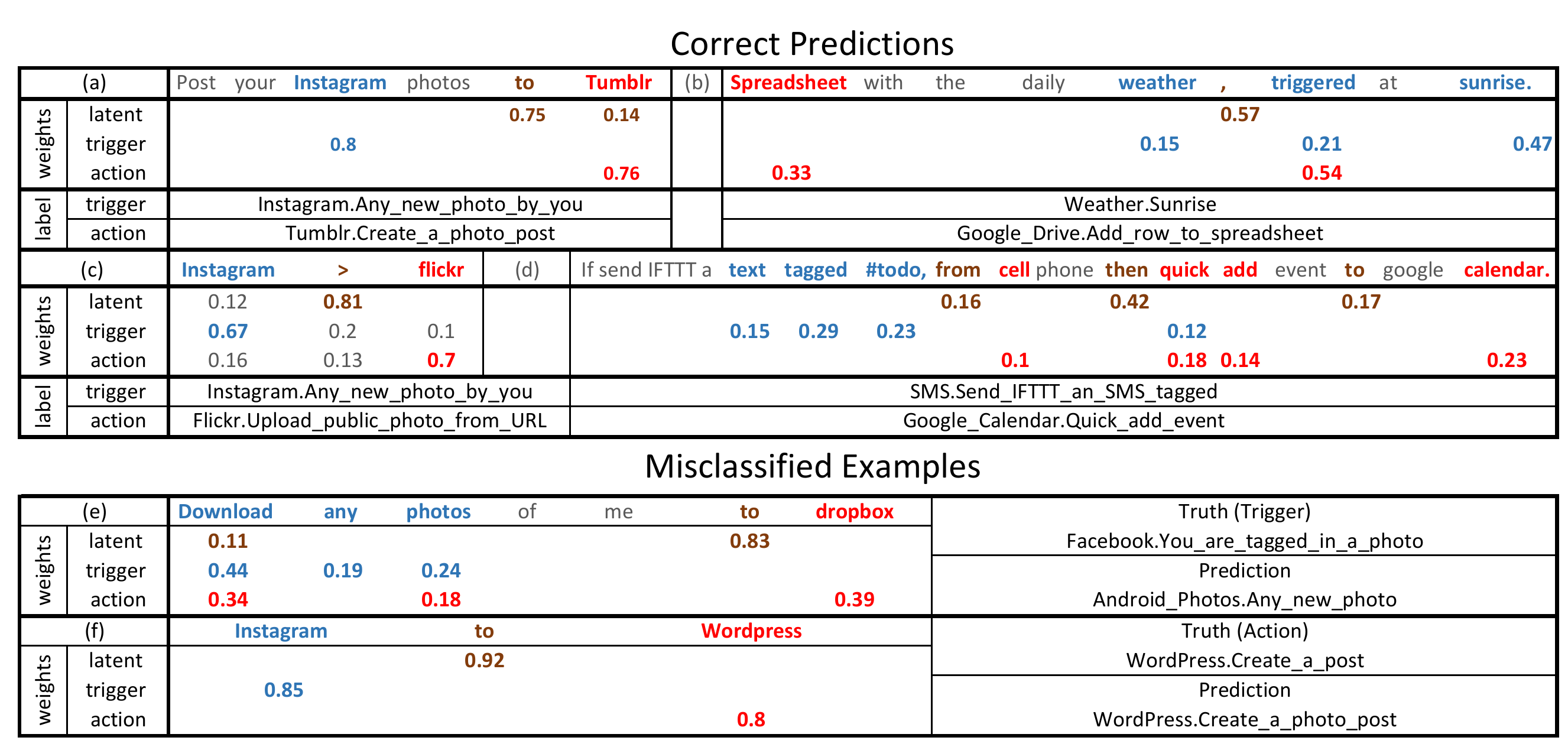}\\
  \caption{Examples of attention weights output by Dict+\dam. \texttt{latent}, \texttt{trigger}, and \texttt{action} indicate the latent weights and
  active weights for the trigger and the action respectively.
  Low values less than $0.1$ are omitted.
  }
  \label{fig:exmp}
\end{figure}

We show some correctly classified and misclassified examples in Figure~\ref{fig:exmp}
along with their attention weights. The weights are computed from a Dict+\dam model.
We choose Dict+\dam instead of BDLSTM+\dam, because the BDLSTM embedding of each
token does not correspond to the token itself only --- it will contain the information passing from
previous and subsequent tokens in the sequence. Therefore, the attention of BDLSTM+\dam is not
as easy to interpret as Dict+\dam.

The latent weights are those used to predict the action functions.
In correctly classified examples, we observe that the latent weights are assigned to
the prepositions that determine which parts of the sentence are associated with the trigger
or the action. An interesting example is (b), where a high latent weight is assigned to ``,".
This indicates that \dam considers ``," as informative as other English words such
as ``to".
We observe the similar phenomenon in
Example (c), where token ``$>$" has the highest latent weight.

In several misclassified examples, we observe that some attention weights may not be assigned correctly.
In Example (e), although there is nowhere explicitly showing
the trigger should be using a Facebook channel, the phrase ``photo of me" hints
that ``me" should be tagged in the photo. Therefore, a human can infer that this should use
a function from the Facebook channel, called ``You\_are\_tagged\_in\_a\_photo". The Dict+\dam
model does not learn this association from the training data. In this example, we expect
that the model should assign high weights onto the phrase ``of me",
but this is not the case, i.e., the weights assigned to ``of" and ``me" are 0.01 and
0.007 respectively. This shows that the Dict+\dam model does not correlate these two
words with the You\_are\_tagged\_in\_a\_photo function. BDLSTM+\dam, on the other
hand, can jointly consider the two tokens, and make the correct prediction.

Example (h) is another example where outside knowledge might help:
Dict+\dam predicts the trigger function to be \texttt{Create\_a\_post} since it does not
learn that Instagram only consists of photos (and low weight was placed on ``Instagram''
when predicting the trigger anyway). Again, BDLSTM+\dam can predict this case correctly.



\paragraph{Acknowledgements.} We thank the anonymous reviewers for their valuable comments.
This material is based upon work partially supported by the National Science Foundation under
Grant No. TWC-1409915, and a DARPA grant FA8750-15-2-0104.
Any opinions, findings, and conclusions or recommendations expressed
in this material are those of the author(s) and do not necessarily
reflect the views of the National Science Foundation and DARPA.

\bibliographystyle{abbrv}
\bibliography{ref}

\begin{thebibliography}{10}

\bibitem{artzi2009broad}
Y.~Artzi.
\newblock Broad-coverage ccg semantic parsing with amr.
\newblock In {\em EMNLP}, 2015.

\bibitem{bahdanau2014neural}
D.~Bahdanau, K.~Cho, and Y.~Bengio.
\newblock Neural machine translation by jointly learning to align and
  translate.
\newblock {\em arXiv preprint arXiv:1409.0473}, 2014.

\bibitem{beltagy-quirk:2016:P16-1}
I.~Beltagy and C.~Quirk.
\newblock Improved semantic parsers for if-then statements.
\newblock In {\em ACL}, 2016.

\bibitem{berant2013semantic}
J.~Berant, A.~Chou, R.~Frostig, and P.~Liang.
\newblock Semantic parsing on freebase from question-answer pairs.
\newblock In {\em EMNLP}, 2013.

\bibitem{branavan2009reinforcement}
S.~R. Branavan, H.~Chen, L.~S. Zettlemoyer, and R.~Barzilay.
\newblock Reinforcement learning for mapping instructions to actions.
\newblock In {\em ACL}, 2009.

\bibitem{chung2014empirical}
J.~Chung, C.~Gulcehre, K.~Cho, and Y.~Bengio.
\newblock Empirical evaluation of gated recurrent neural networks on sequence
  modeling.
\newblock {\em arXiv preprint arXiv:1412.3555}, 2014.

\bibitem{dong2016language}
L.~Dong and M.~Lapata.
\newblock Language to logical form with neural attention.
\newblock In {\em ACL}, 2016.

\bibitem{gulwani2014nlyze}
S.~Gulwani and M.~Marron.
\newblock Nlyze: Interactive programming by natural language for spreadsheet
  data analysis and manipulation.
\newblock In {\em SIGMOD}, 2014.

\bibitem{jones2012semantic}
B.~K. Jones, M.~Johnson, and S.~Goldwater.
\newblock Semantic parsing with bayesian tree transducers.
\newblock In {\em ACL}, 2012.

\bibitem{kate2005learning}
R.~J. Kate, Y.~W. Wong, and R.~J. Mooney.
\newblock Learning to transform natural to formal languages.
\newblock In {\em AAAI}, 2005.

\bibitem{kingma2014adam}
D.~Kingma and J.~Ba.
\newblock Adam: A method for stochastic optimization.
\newblock {\em arXiv preprint arXiv:1412.6980}, 2014.

\bibitem{kushman2013using}
N.~Kushman and R.~Barzilay.
\newblock Using semantic unification to generate regular expressions from
  natural language.
\newblock In {\em NAACL}, 2013.

\bibitem{le2013smartsynth}
V.~Le, S.~Gulwani, and Z.~Su.
\newblock Smartsynth: Synthesizing smartphone automation scripts from natural
  language.
\newblock In {\em MobiSys}, 2013.

\bibitem{lei2013natural}
T.~Lei, F.~Long, R.~Barzilay, and M.~C. Rinard.
\newblock From natural language specifications to program input parsers.
\newblock In {\em ACL}, 2013.

\bibitem{DBLP:journals/corr/LingGHKSWB16}
W.~Ling, E.~Grefenstette, K.~M. Hermann, T.~Kocisk{\'{y}}, A.~Senior, F.~Wang,
  and P.~Blunsom.
\newblock Latent predictor networks for code generation.
\newblock {\em CoRR}, 2016.

\bibitem{quirk2015language}
C.~Quirk, R.~Mooney, and M.~Galley.
\newblock Language to code: Learning semantic parsers for if-this-then-that
  recipes.
\newblock In {\em ACL}, 2015.

\bibitem{memn2n}
S.~Sukhbaatar, J.~Weston, R.~Fergus, et~al.
\newblock End-to-end memory networks.
\newblock In {\em NIPS}, 2015.

\bibitem{vinyals2015grammar}
O.~Vinyals, {\L}.~Kaiser, T.~Koo, S.~Petrov, I.~Sutskever, and G.~Hinton.
\newblock Grammar as a foreign language.
\newblock In {\em NIPS}, 2015.

\bibitem{wong2006learning}
Y.~W. Wong and R.~J. Mooney.
\newblock Learning for semantic parsing with statistical machine translation.
\newblock In {\em NAACL}, 2006.

\bibitem{xu2015show}
K.~Xu, J.~Ba, R.~Kiros, A.~Courville, R.~Salakhutdinov, R.~Zemel, and
  Y.~Bengio.
\newblock Show, attend and tell: Neural image caption generation with visual
  attention.
\newblock {\em arXiv preprint arXiv:1502.03044}, 2015.

\bibitem{zaremba2014recurrent}
W.~Zaremba, I.~Sutskever, and O.~Vinyals.
\newblock Recurrent neural network regularization.
\newblock {\em arXiv preprint arXiv:1409.2329}, 2014.

\bibitem{zelle1996learning}
J.~M. Zelle.
\newblock Learning to parse database queries using inductive logic programming.
\newblock In {\em AAAI}, 1996.

\end{thebibliography}

\newpage

\ifarxiv
\appendix

\begin{figure}[t]
	\centering
	\includegraphics[scale=0.45]{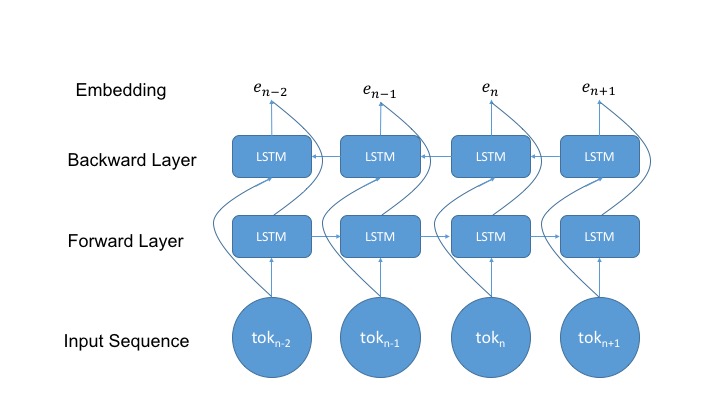}\\
	\caption{BDLSTM Embedding}\label{fig:bdlstm}
\end{figure}

\section{BDLSTM and attention model details}
\label{app:arch}

\subsection{BDLSTM embedding}
Recurrent neural networks have become popular for natural language processing tasks due to their suitability for processing sequential data.
Given inputs $\mathbf{x}_1$ to $\mathbf{x}_J \in \mathbb{R}^n$, a RNN computes
\[ \mathbf{h}_t = \tanh(\mathbf{W}_{xh} \mathbf{x}_t + \mathbf{W}_{hh} \mathbf{h}_{t-1} + \mathbf{b}_h) \]
where $\mathbf{h}_0$ is a zero vector, $\mathbf{W}_{xh}$ and $\mathbf{W}_{hh}$ are trained parameter matrices
respectively of size $m \times n$ and $n \times n$, and $\mathbf{b}_h \in \mathbf{R}^m$ is used as a bias.

Long Short-Term Memory (LSTM) is a RNN variant which is better suited for learning long-term dependencies.
Although several versions of it have been described in the literature, we use the version in Zaremba et al. \cite{zaremba2014recurrent} and borrow their notation here:
\begin{align*}
\begin{pmatrix}
i \\ f \\ o \\ g
\end{pmatrix}
&=
\begin{pmatrix}
\sigma \\
\sigma \\
\sigma \\
\text{tanh}
\end{pmatrix}
\mathbf{T}_{2n,4n}
\begin{pmatrix}
\mathbf{x}_t \\
\mathbf{h}_{t-1}
\end{pmatrix} \\
\mathbf{c}_t &= f \odot \mathbf{c}_{t-1} + i \odot g \\
\mathbf{h}_t &= o \odot \tanh(c_t)
\end{align*}
Here, $\sigma$ is the sigmoid function, and $\odot$ denotes the element-wise multiplication. The \emph{memory cells} $\mathbf{c}_t$ are designed to store information for longer periods of time
than the hidden state.

We construct the bi-directional model with
a forward LSTM which receives the input sequence in the original order, and
a backward LSTM which receives the input sequence in the reverse order. The
BDLSTM embedding is the concatenation of the output of the two. This structure is illustrated in Figure~\ref{fig:bdlstm}.

\subsection{Standard attention model}

The standard attention model differs with \NewAttention in the way that there is only one layer of active attention. In particular, we have

\paragraph{The attention layer.}
We compute the attention $a$ over the $J$ tokens with the following:
\[a = \mathbf{softmax}(u^T\emb{X}{\theta_1}).\]
$a$ has $J$ dimensions and $u$ is a $d$-dimensional trainable vector.

\paragraph{Output representation.} We use a third set of parameters $\theta_3$ to embed $X$, and then the final output, a $d$-dimension vector, is the
weighted-sum of these embeddings using the active weights.
\[o=\emb{X}{\theta_2}a\]

\paragraph{Prediction.} We compute probabilities over the output class labels by a matrix multiplication followed by softmax: \[\hat{f}=\mathbf{softmax}(Wo)\]

\begin{figure}[t]
	\centering
	\includegraphics[scale=0.25]{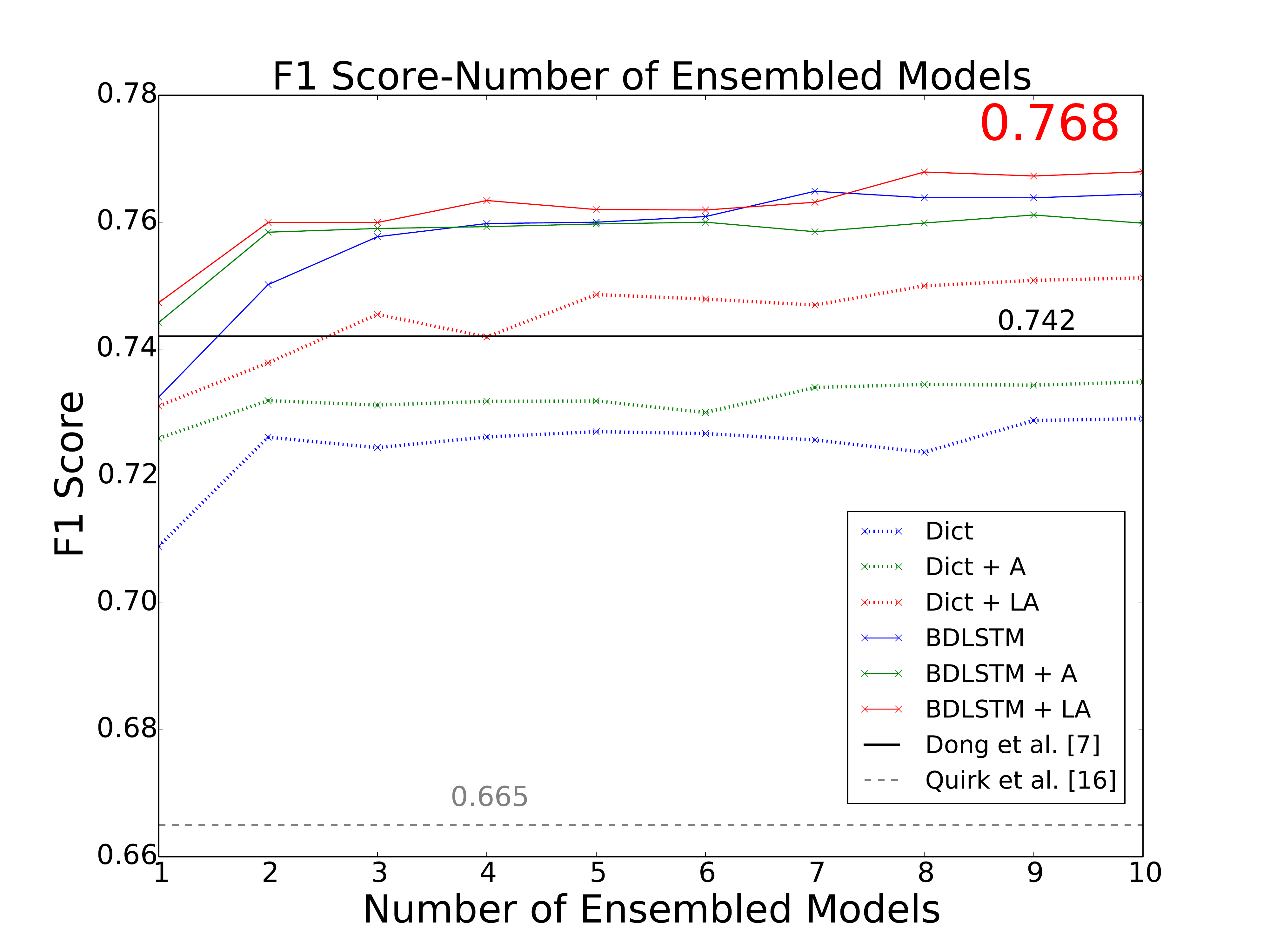}
	\caption{$F_1$ score for arguments prediction}
	\label{fig:f1-arg}
\end{figure}

\section{Predicting Arguments}
\label{app:arg}

We provide a frequency-based method for predicting the function arguments as a baseline, and show that this can outperform existing approaches dramatically when combined with our higher-performance function name prediction. In particular, for each description, we first predict
the (trigger and action) functions $f_t, f_a$. For each function $f$, for each
argument $a$, and for each possible argument value $v$, we compute the frequency
that $f$'s argument $a$ takes the value $v$. We denote this frequency as $\mathit{Pr}(v|f, a)$. Our prediction is made by computing
\[\mathrm{argmax}_v \mathit{Pr}(v|f, a).\]
Note that the prediction is made entirely based on the predicted function $f$,
without using any information from the description.

\newcommand{\missingtok}{$\langle MISSING \rangle$}

We found that for a given function, some arguments may not
appear in all recipes using this function. In this case, we give the
value a special token, \missingtok; this is distinct from the case where the
argument exists but its value has zero length (i.e., ``").

We use the same setup as in Section~\ref{sec:exp}. The results are presented in Figure~\ref{fig:f1-arg}. \cite{beltagy-quirk:2016:P16-1} does not present
their results for arguments prediction, so we do not include it in Figure~\ref{fig:f1-arg}.
We can observe that the results are basically consistent with the results for channel and
function accuracy. 

\section{Data statistics and numerical results}
\label{app:data:stat}

\begin{table}[t]
\begin{centering}
\begin{tabular}{|c|c|c|}
\hline
& Training & Test (Gold)\\
\hline
\# of trigger channels  & 112 & 59 \\
\# of trigger functions  & 443 & 136\\
\# of action channels  & 87 & 41\\
\# of action functions  & 161 & 56 \\
\hline
\# of recipes &  68,083 & 584\\
\hline
\end{tabular}
\par\end{centering}
\caption{Statistics for IFTTT dataset}
\label{tab:data:stat}
\vspace{-0.5cm}
\end{table}

\begin{table}[t]
\begin{tabular}{c}
	Channel Accuracy for Ensembled Models (Fig.~\ref{fig:acc-channel})
	\\
	\begin{tabular}{|l|cccccccccc|}
		\hline
		Ensemble & 1 & 2 & 3 & 4 & 5 & 6 & 7 & 8 & 9 & 10\\
		\hline
Dict & 71.9 & 72.8 & 73.5 & 74.1 & 74.7 & 80.1 & 80.5 & 79.6 & 80.5 & 81.3\\
Dict+A & 82.4 & 83.0 & 83.6 & 83.2 & 83.2 & 83.2 & 83.7 & 83.9 & 83.6 & 83.7\\
Dict+LA & 87.3 & 87.7 & 88.5 & 87.7 & 87.7 & 87.3 & 87.0 & 86.4 & 86.4 & 87.5\\
BDLSTM & 84.8 & 89.2 & 90.1 & 90.4 & 90.6 & 90.8 & 90.4 & 90.9 & 91.4 & 91.6\\
BDLSTM+A & 89.2 & 90.4 & 90.4 & 89.7 & 90.4 & 90.4 & 90.8 & 90.9 & 90.8 & 91.1\\
BDLSTM+LA & 89.6 & 89.9 & 90.2 & 90.4 & 90.8 & 90.8 & 90.9 & 90.9 & 91.4 & 91.6\\
\hline
Dong et al. [3] & \multicolumn{10}{c|}{81.4}\\
\hline
Beltagy et al. [7] & \multicolumn{10}{c|}{89.7}\\
\hline
Quirk et al. [16] & \multicolumn{10}{c|}{89.1}\\
\hline
	\end{tabular}
  \vspace{0.2cm} \\ 
	Function Accuracy for Ensembled Models (Fig.~\ref{fig:acc-func})
	\\
	\begin{tabular}{|l|cccccccccc|}
		\hline
		Ensemble & 1 & 2 & 3 & 4 & 5 & 6 & 7 & 8 & 9 & 10\\
		\hline
Dict & 71.6 & 74.7 & 74.7 & 75.9 & 76.0 & 76.0 & 75.7 & 75.7 & 76.0 & 76.4\\
Dict+A & 74.0 & 76.0 & 75.9 & 76.0 & 76.4 & 75.7 & 76.5 & 77.6 & 77.2 & 77.2\\
Dict+LA & 79.6 & 78.4 & 78.0 & 78.9 & 78.0 & 79.9 & 79.9 & 79.9 & 81.3 & 82.2\\
BDLSTM & 78.6 & 81.8 & 81.5 & 82.4 & 84.1 & 85.4 & 85.6 & 86.0 & 85.8 & 85.4\\
BDLSTM+A & 80.3 & 83.6 & 84.6 & 84.4 & 84.6 & 84.4 & 84.4 & 84.6 & 85.1 & 84.8\\
BDLSTM+LA & 82.4 & 83.7 & 85.3 & 86.0 & 85.8 & 85.6 & 86.0 & 86.8 & 87.5 & 87.3\\
\hline
Dong et al. [3] & \multicolumn{10}{c|}{78.4}\\
\hline
Beltagy et al. [7] & \multicolumn{10}{c|}{82.5}\\
\hline
Quirk et al. [16] & \multicolumn{10}{c|}{71.0}\\
		\hline
 	\end{tabular}
  \vspace{0.2cm} \\
 	F1 Score for Arguments for Ensembled Models (Fig.~\ref{fig:f1-arg})
 	\\
	\begin{tabular}{|l|cccccccccc|}
		\hline
		Ensemble & 1 & 2 & 3 & 4 & 5 & 6 & 7 & 8 & 9 & 10\\
		\hline
Dict & 70.9 & 72.6 & 72.4 & 72.6 & 72.7 & 72.7 & 72.6 & 72.4 & 72.9 & 72.9\\
Dict+A & 72.6 & 73.2 & 73.1 & 73.2 & 73.2 & 73.0 & 73.4 & 73.4 & 73.4 & 73.5\\
Dict+LA & 73.1 & 73.8 & 74.5 & 74.2 & 74.9 & 74.8 & 74.7 & 75.0 & 75.1 & 75.1\\
BDLSTM & 73.2 & 75.0 & 75.8 & 76.0 & 76.0 & 76.1 & 76.5 & 76.4 & 76.4 & 76.4\\
BDLSTM+A & 74.4 & 75.8 & 75.9 & 75.9 & 76.0 & 76.0 & 75.8 & 76.0 & 76.1 & 76.0\\
BDLSTM+LA & 74.7 & 76.0 & 76.0 & 76.3 & 76.2 & 76.2 & 76.3 & 76.8 & 76.7 & 76.8\\
\hline
Dong et al. [3] & \multicolumn{10}{c|}{74.2}\\
\hline
Quirk et al. [16] & \multicolumn{10}{c|}{66.5}\\
		\hline
	\end{tabular}
\end{tabular}
\caption{Numerical Results for Figure~\ref{fig:acc-channel}~\ref{fig:acc-func}, and~\ref{fig:f1-arg}}
\label{tab:num-res}
\end{table}

In this section, we provide concrete data statistics and results. The statistics for
IFTTT dataset that we evaluated is presented in Table~\ref{tab:data:stat}. The
numerical values corresponding to
Figure~\ref{fig:acc-channel},~\ref{fig:acc-func}, and~\ref{fig:f1-arg} are
presented in Table~\ref{tab:num-res}. The statistics for the data used in one-shot learning are presented in Table~\ref{tab:exp3:stat}. The numerical results
corresponding to Figure~\ref{fig:top100} and~\ref{fig:nontop100} are presented in Table~\ref{tab:num-res-one-shot}.

\begin{table}[t!]
\centering
\begin{tabular}{|c|c|c|}
\hline
& SkewTop100 & SkewNonTop100\\
\hline
\# of recipes & 61,341 & 10,707\\
\# of recipes in $S$ & 58,376  & 9,707 \\
\# of recipes not in $S$ & 2,965 & 1,000\\
\hline
\end{tabular}
\caption{Statistics for unbalanced training sets\vspace{-0.5cm}}
\label{tab:exp3:stat}
\end{table}
\begin{table}[t!]
	\includegraphics[scale=0.8]{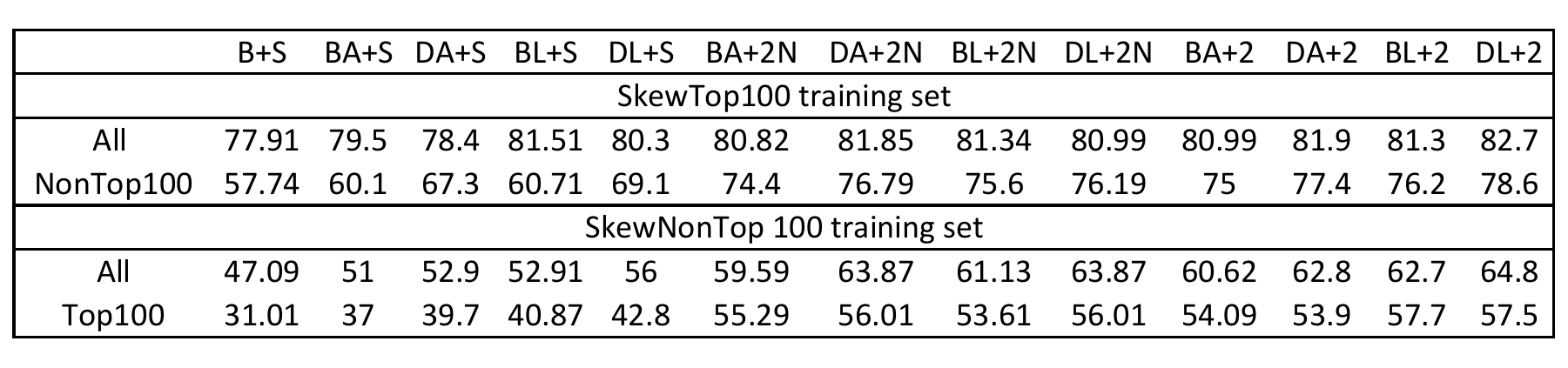}
    \caption{Numerical Results For Figure~\ref{fig:top100} and~\ref{fig:nontop100}}
    \label{tab:num-res-one-shot}
\end{table}



\fi

\end{document}